\documentclass{article}

\usepackage{arxiv}

\usepackage[utf8]{inputenc} 
\usepackage[T1]{fontenc}    
\usepackage{hyperref}       
\usepackage{url}            
\usepackage{booktabs}       
\usepackage{amsfonts}       
\usepackage{nicefrac}       
\usepackage{microtype}      
\usepackage{lipsum}
\usepackage{graphicx}
\usepackage{comment}
\usepackage{amsmath,amssymb} 
\usepackage{color}
\usepackage{bm}
\usepackage{microtype}
\usepackage{colortbl}
\usepackage[dvipsnames]{xcolor}
\usepackage{pifont}
\usepackage{enumitem}

\newcommand{\cmark}{\ding{51}}%
\newcommand{\xmark}{\ding{55}}%
\newcommand{\myparagraph}[1]{\textbf{#1} --- }

\usepackage{dirtytalk}

\graphicspath{ {./images/} }

\title{Estimating semantic structure for the VQA answer space}

\author{
 Corentin Kervadec \\
  Orange Labs, Cesson-Sévigné, France\\
  Université de Lyon, INSA-Lyon, LIRIS\\
  UMR CNRS 5205, Villeurbanne, France\\
  \texttt{corentin.kervadec@orange.com} \\
   \And
 Grigory Antipov \\
  Orange Labs,\\
  Cesson-Sévigné, France\\
  \texttt{grigory.antipov@orange.com} \\
    \And
 Moez Baccouche \\
  Orange Labs,\\
  Cesson-Sévigné, France\\
  \texttt{moez.baccouche@orange.com} \\
    \And
 Christian Wolf \\
  Université de Lyon, INSA-Lyon, LIRIS\\
  UMR CNRS 5205, Villeurbanne, France\\
  \texttt{christian.wolf@insa-lyon.fr} \\
}

\begin{document}
\maketitle

\begin{abstract}

Since its appearance, Visual Question Answering (VQA, i.e. answering a question posed over an image), has always been treated as a classification problem over a set of predefined answers. Despite its convenience, this classification approach poorly reflects the semantics of the problem limiting the answering to a choice between independent proposals, without taking into account the similarity between them (e.g. equally penalizing for answering \say{cat} or \say{German shepherd} instead of \say{dog}). We address this issue by proposing (1) two measures of proximity between VQA classes, and (2) a corresponding loss which takes into account the estimated proximity. This significantly improves the generalization of VQA models by reducing their language bias. In particular, we show that our approach is completely model-agnostic since it allows consistent improvements with three different VQA models. Finally, by combining our method with a language bias reduction approach, we report SOTA-level performance on the challenging VQAv2-CP dataset.

\end{abstract}

\section{Introduction}
Visual Question Answering (VQA) is a task which requires to provide a textual answer given a question and an image as input. When properly formulated, this problem requires a high-level understanding of the content of the image as well as the problem statement (the question), and therefore is often considered to be a proxy task allowing to evaluate the visual reasoning abilities of a
system.

While the problem itself requires the prediction of textual output (the answer word or sentence), which is an output space with rich structure, most, if not all known benchmarks and evaluation prototocols deal with it as a classification problem, for instance VQAv1~\cite{antol2015vqa}, VQAv2~\cite{goyal2017making}, GQA~\cite{hudson2019gqa}. This is done by creating a dictionary of output answer classes constructed from the most frequent answers of the training set.
Models addressing the problem are asked to predict a probability distribution over this dictionary and are usually trained with  cross-entropy loss plus eventual self-supervised auxiliary losses.

The merit of this approach is its ease of use, its straightforward definition of loss function and its empirical success. It, however, raises two fundamental issues: 
(i) the dictionary highly depends on the training set and hurts generalization to unseen data;
(ii) the answer classes of the dictionary are considered to be independent, without taking into account their semantic relationships.
This results in models highly dependant on question biases, as observed in \cite{vqa-cp}.

\begin{figure}[t]
  \begin{center}
     \includegraphics[width=1.0\linewidth]{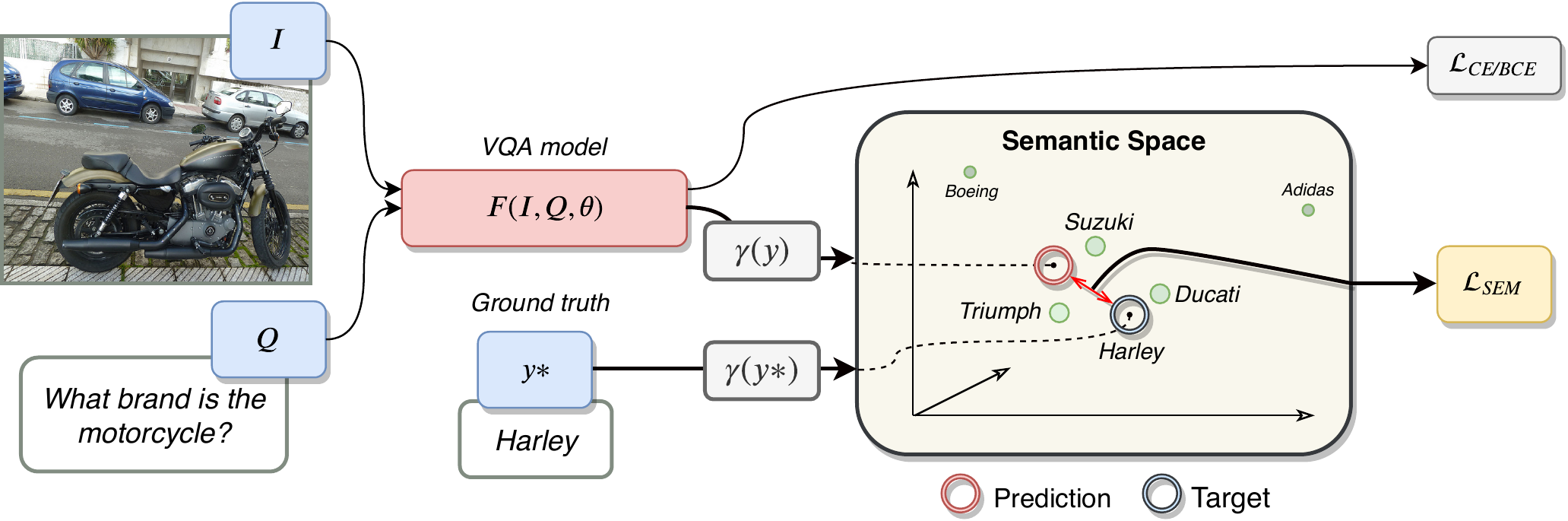}
  \end{center}
     \caption{We introduce a new loss structuring the semantic space of VQA output classes. The prediction $\bm{y}$ and the target label $\bm{y}^*$ are both projected into a semantic space using $\gamma(\cdot)$. Then, the model is trained to minimize the distance between the prediction and the target in that space using our semantic loss $\mathcal{L}_{\textit{SEM}}$.}
  \label{fig:general}
\end{figure}

While the long-term objective of the community is arguably to move to a direct structured prediction of the textual output, in this work we focus on the second issue (ii), arguing that properly structuring the semantic space of output classes can overcome some of the shortcomings of the classification strategy.
We address this through a new loss function (that we call \textit{semantic loss}), which measures the semantic distance between the prediction and the ground truth answer. If, for instance, the question is  ``\emph{Who is on the car?}'' and the expected answer is ``\emph{woman}'', we propose to penalize the wrong answer ``\emph{girl}'' less than an also wrong answer ``\emph{boy}'', while the classical cross-entropy loss would penalize both wrong answers equally (see Figure~\ref{fig:general}).

We show that this new loss provides two different benefits: (i) the more direct benefit of pushing error cases to more favorable answer classes, like the ``\emph{girl}'' instead of ``\emph{boy}'' example above. This alone should provide a quality improvement in many applications, albeit not measured through the standard metrics \emph{Accuracy, Recall, Precision}, not adapted to measure the reasoning abilities of an agent; (ii) we also show that structuring the output space improves performance in absolute terms as measured through \emph{Accuracy}, therefore drives the model to make fewer errors whatever their type might be.

While the intuitive notion of semantic proximity is easy to grasp for a human, its formal definition is less clear. We propose two different methods to estimate the proximity of two answer words or sentences from data: the first one exploits the relationship between answer class and answer text and makes use of classical self-supervised pre-training of word embeddings, while the second one extracts statistics from the multiple redundant ground truth annotations of VQA datasets.

During training, we combine the new loss along with the standard cross-entropy discriminative loss. In the experiments we show that this improves accuracy on the VQA task and, in particular, helps reducing the dependency on language biases. More so, we show that the performance improvements obtained by our contribution are complementary to other efforts in removing language bias \cite{cadene2019rubi}, providing strong combined performance. We believe this further indicates the interest of this new learning signal.

Our contributions are as follows:
\begin{enumerate}
  \item We design a semantic loss for VQA, which helps the model to better structure its  answer space and to learn the semantic relations between the answers.
  \item We propose two different methods for estimating semantic proximity between  answer classes based on word embeddings and annotation statistics, respectively.
  \item We demonstrate the effectiveness of our method on the VQAv2-CP~\cite{vqa-cp} and VQAv2~\cite{goyal2017making} datasets with three different neural models, showing consistency of the improvement over datasets and models. 
  \item When combined with other efforts in addressing language bias, our performance improvements add up and achieve performance on-par with the State-Of-The-Art (SOTA) on VQAv2-CP with reasonably complex model architectures.
\end{enumerate}

\section{Related work}

\myparagraph{VQA as a classification task}
The broad majority of works are approaching VQA as a classification task. This strategy simplifies the supervision part of the training and makes the VQA approaches easily comparable between each other. Indeed, many VQA approaches, including attention-based networks~\cite{yang2016stacked}, object-based attention mechanisms~\cite{anderson2018bottom}, bilinear fusion methods~\cite{kim2018bilinear}, and, more recently, Transformer~\cite{vaswani2017attention}-based models~\cite{yu2019deep}, have been introduced and evaluated on the VQA classification benchmarks.

\myparagraph{Biases in VQA datasets}
The success of these works is in part due to the creation of large annotated VQA corpus. The VQAv1~\cite{antol2015vqa} dataset gathers more than 200k real-world images annotated with 760K questions in natural language. Moreover, each question is annotated with 10 ground truth answers in order to model the ambiguity and the disagreement between annotators. Nevertheless, as the data collection is fastidious and costly, VQAv1 suffers from numerous language biases as pointed out by many works such as~\cite{goyal2017making}, and ~\cite{agrawal2016analyzing}.
In order to downplay the influence of the language bias, \cite{goyal2017making} released an updated version of the same dataset (baptized VQAv2) by carefully balancing the answer distribution per question type.
\cite{johnson2017clevr} went one step further by designing a fully synthetic dataset called CLEVR in which automatically generated questions are asked on synthetically generated 3D images. This corpus is thought as a diagnostic dataset aimed at evaluating the visual reasoning capability of the models.
However, the limited environment of CLEVR~\cite{johnson2017clevr} prevents from generalizing to complex realistic images such as in VQAv2~\cite{goyal2017making}.
Finally,~\cite{hudson2019gqa} built the GQA dataset, a semi-synthetic dataset where automatically generated questions are asked on real images, which can be seen as an intermediate step between CLEVR and VQAv2.

\myparagraph{The generalization curse of VQA}
Nevertheless, despite the efforts made on data collection, the language bias issue persists and VQA models continue to suffer from the generalization curse. Early works raised an alarm by diagnosing many drawbacks on VQA models, such as their tendency to pay little attention to the image and to \say{only read half of the question}~\cite{agrawal2016analyzing}. Similarly,~\cite{vqahat} showed that attention-based models do not attend to the same visual regions as humans do. More recently,~\cite{hendricks2018women} pointed out the gender biases learned by image captioning models. To better diagnose the generalization gap on VQA,~\cite{vqa-cp} reorganized VQAv1~\cite{antol2015vqa} and VQAv2~\cite{goyal2017making} into the VQA-CP (VQA under Changing Priors), a new dataset where the per-question answer distribution of the train split is made explicitly different from the one in the test split. In particular, they show that a blind model (which only has access to the question without seeing the image) achieves a surprisingly high accuracy on VQAv2 -- $44\%$ -- whereas it reaches only $16\%$ on VQAv2-CP~\cite{vqa-cp}. Moreover, many of the successful models on VQA datasets fall short on VQA-CP unveiling their lack in visual understanding and their tendency to rely on question biases. 

\myparagraph{Reducing language biases on VQA}
Therefore, several works have recently tried to tackle this generalisation issue. \cite{ramakrishnan2018overcoming} trained their model using an adversarial game against a question-only adversary in order to discourage the base model to rely on language prior. Similarly, the authors of RUBi~\cite{cadene2019rubi} added a question-only branch to the base model during the training to adapt its prediction in order to prevent it from learning question biases. Other methods make use of additional annotated supervision to improve the generalization capability. Using the VQA-HAT dataset's annotations~\cite{vqahat}, the HINT~\cite{selvaraju2019taking} model is supervised to attend to the same visual regions as humans.
\cite{wu2019self} built upon HINT proposing even a more sophisticated approach by carefully designing a three-steps learning strategy, named Self-Critical Reasoning (SRC).
SRC accentuates the model sensitivity to the important visual regions.
It should be noted that SRC requires additional data annotations such as human attention maps~\cite{vqahat} or textual explanations~\cite{huk2018multimodal}.
Finally,~\cite{jingovercoming} introduced a Decomposed Linguistic Representation\footnote{To the best of our knowledge, at the time of writing of our paper, paper \cite{jingovercoming} has not yet been accepted to a peer-reviewed conference or journal.} (DLR) approach consisting in learning to decompose the question into the \say{type representation}, \say{object representation} and \say{concept representation}. Although it allows to improve the model's accuracy on VQAv2-CP~\cite{vqa-cp}, it causes a significant drop of performance on VQAv2~\cite{goyal2017making}. 
In this paper, we contribute to these efforts as we demonstrate how a semantic loss, helping the model to structure its output space, permits to reduce dependency towards language biases.
 
\paragraph{}
Our method has also connections with distributed encoding approaches which have been successfully applied on age estimation from faces~\cite{geng2013facial} and more generally on label distribution learning \cite{geng2016label}.
Additionally, an early VQA work, the DAQUAR~\cite{malinowski2014multi} dataset, pioneers the use of soft evaluation for VQA. They use a variant of the Wu Palmer similarity~\cite{wu1994verbs} over a lexical database to compute a soft prediction score. This way, a prediction semantically close to the target answer is no longer considered as false. This allows to obtain finer evaluation of performances on VQA. However, it is only used for evaluation and such metric has fundamental drawbacks, such as its inability to discriminate colors.

\section{Structuring the answer space}

\noindent
As our contribution is agnostic w.r.t. particular model architectures, we consider a VQA model as a whitebox function $f$ taking into account an image $\bm{I}$, a question $\bm{q}$ and producing an answer $\bm{y}$, which is an output vector over the output alphabet:
\begin{equation}
  \label{eq:vqa}
  \bm{y} = \textit{f}\left(\bm{I}, \bm{q}; \theta\right)\textit{.}
\end{equation}
In the literature, the cross-entropy loss is frequently used to measure the prediction error during training, which casts the task as a classification problem with a single unique correct answer per problem instance:
\begin{align}
  \label{eq:ce}
  \mathcal{L}_{\textit{CE}} &= \textit{CE}\left(\textit{softmax}\left(\bm{y}\right), \bm{y}^*\right)\textit{,} \\
  & = \sum_i^{N_c} \bm{y}^*_i \log(\textit{softmax}\left(\bm{y}\right)_i)
\end{align}
where $\textit{CE}(\cdot,\cdot)$ is the cross-entropy function, $N_c$ is the size of the answer dictionary (hence the number of classes), and $\bm{y}^*$ is the one-hot encoded vector of ground truth answers.
Alternatively, some datasets (such as VQAv1~\cite{antol2015vqa} and VQAv2~\cite{goyal2017making}) admit more than one correct answer for a given question. This allows to take into account ambiguities in the question formulations and in the annotation uncertainties. In that case, an appropriate formulation is a model allowing to predict more than one answer class combined with a soft binary cross-entropy loss given as:
\begin{align}
  \label{eq:bce}
  \mathcal{L}_{\textit{BCE}} &= \frac{1}{N_c} \sum_i^{N_c} \textit{BCE}\left(\textit{sigmoid}\left(\bm{y}_i\right), \bm{y}^*_i\right)\textit{,}\\
  & = \frac{1}{N_c} \sum_i^{N_c} \bm{y}^*_i . \log\left(\textit{sigmoid}(\bm{y}_i)\right) + (1-\bm{y}^*_i) . \log\left( 1 - \textit{sigmoid}(\bm{y}_i) \right)
\end{align}
where $\textit{BCE}(\cdot,\cdot)$ is the
 binary cross-entropy, $\bm{y}_i$ and $\bm{y}^*_i$ are respectively the predicted and the ground truth probability of
 answer $i$ of the dictionary. 

$\mathcal{L}_{\textit{CE}}$ and $\mathcal{L}_{\textit{BCE}}$ are, both, widely used learning signals~\cite{anderson2018bottom,goyal2017making,yu2019deep,cadene2019rubi,kim2018bilinear}, and despite their difference, they share the common shortcoming of not taking into account eventual differences in the semantic proximity of answers.

To address this issue, in this work, we introduce a new semantic loss, which provides additional structure to the output space.
Defining the semantic loss requires to set up (i) a semantic space which embeds the answers,
and (ii) a distance function measuring the semantic similarity between two answers.


\vspace{2mm}

\noindent
\myparagraph{Projection to a semantic space}
\label{subsection:embed}
a semantic space is required to satisfy several proprieties. It needs:
\begin{enumerate}[label=\alph*]
\item - to be structured, i.e. to take into account semantic proximity estimated from some data source, \label{point:semantic}
\item - to be able to cope with the continuous nature of neural networks, which provide continuous estimates for each output class (estimates of posterior probabilities if trained with cross-entropy) as opposed to discrete symbolic predictions. \label{point:continuous}
\end{enumerate}

\noindent
We address requirement (\ref{point:continuous})  by defining a function $\gamma$, which projects the continuous prediction and the discrete ground truth label, respectively, into a joint semantic space:
\begin{align}
  \gamma(\bm{y}) &= 
\sum_{i \in top_{k}(\bm{y})}^{N_c} \bm{y}_i g(i)\textit{,} \label{eq:embed}
\end{align}
The sum in (\ref{eq:embed}) weights the different class mappings $g$ by their predicted output, and is defined over the top-$k$ highest predictions $\bm{y}_i$, where $k$ is a hyper-parameter. This allows the mapping to be dominated by the highest-probably answer classes and eliminate spurious non-probable influences.

The function $\gamma$ depends on the mapping $g$, which needs to address requirement (\ref{point:semantic}). We propose two different semantic spaces, which estimate the output space structure from two different data sources, namely, respectively, pre-trained word embeddings and redundancies in VQA dataset groundtruth annnotations. These two learning signals have fundamentally different origins.

\begin{description}
  \item[Glove] --- Word embeddings are widely-used projections from a discrete symbol space to a continuous vectorial space \cite{pennington2014glove,mikolov2013distributed}. They are trained in a self-supervised way, minimizing the error in predicting words from a context, i.e. groups of words around the predicted target. In other words, the semantic space emerges as a side product of the estimation of co-occurrences in language space. The  Glove representation~\cite{pennington2014glove} has been shown to capture fine-grained semantic and syntactic regularities, and as such is thus a natural choice for our semantic space.
  We use the GloVe embedding directly to project output class $i$ to its vectorial representation:
  \begin{equation}
    g(i) = \textrm{Glove}
    \left (
    a_i \right )
  \end{equation}
  where $a_i$ is the textural representation of answer the class $i$.
  
  \vspace{3mm}
  
  \item[Co-oc] --- While the word embeddings mentioned beforehand exploit statistical regularities in large text corpus to estimate semantic proximity, we propose an alternative which directly taps into human assessments. Annotations on semantic distances are hard to come by, so we provide estimates from alternative human annotations. The two datasets VQAv2~\cite{goyal2017making} and VQAv2-CP~\cite{vqa-cp} provide 10 ground truth answers per question obtained by 10 different people. We define the semantic proximity of a pair of answers classes $i$ and $j$ as the amount of coherence between these two classes in terms of human annotations. To be more precise, we estimate it from the co-occurrences $\bm{c}_{ij}$ of the two answer classes in different question instances:
  
  \begin{equation}
    \bm{c}_{ij} = \frac{
    \log \left | (i,j) \right |}
    {
    \log \left |(i) \right |
    \log \left |(j) \right |} 
  \end{equation}
  where $(i)$ and $(j)$, are the sets of question instances where answers $i$ and $j$, respectively, occurs, and $(i,j)$ is the set of question instances where both answers $i$ and $j$ occurs.
  The Co-oc embedding vector is then defined as follows:
  \begin{equation}
    g(i) =  \left [ \bm{c}_{i1},\dots, \bm{c}_{ij}, \dots, \bm{c}_{iN_c}
    \label{eq:cooc}
    \right ]
  \end{equation}
  In Co-oc space, two answers are close if they are likely to be used as answers for a same question. 
\end{description}

\begin{figure}[t]
  \begin{center}
     \includegraphics[width=1.0\linewidth]{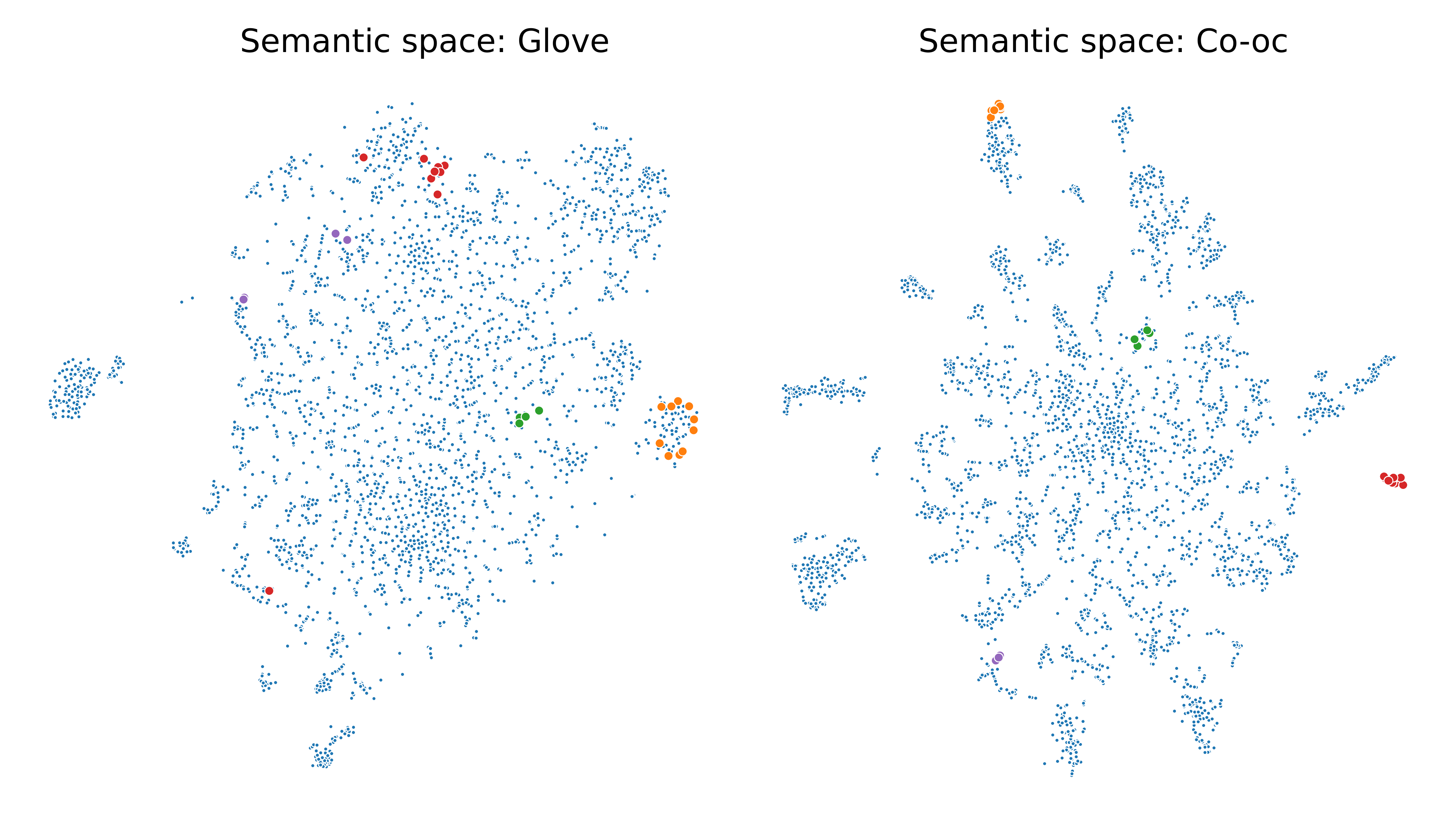} \\
     {\small 
     \begin{tabular}{ccc}
          \textbf{Category} & \textbf{Color} &  \textbf{VQA answer classes}\\
          \hline
          {\color{orange}Colors} & {\color{orange}Orange} & {\color{orange}orange, white, red, blue, green, gray, black, pink, brown, yellow}\\
          {\color{red}Dogs} & {\color{red}Red} & {\color{red}puppy, golden retriever, german shepherd, husky, terrier} \\
          && {\color{red}labrador, sheepdog, rottweiler, corgi}
          \\
          {\color{Fuchsia}Motorcyles} & {\color{Fuchsia}Purple} & {\color{Fuchsia}yamaha, kawasaki, harley, suzuki} \\
          {\color{OliveGreen}Trees} & {\color{OliveGreen}Green}  & {\color{OliveGreen}log, palm tree, tree branch, christmas tree} \\
     \end{tabular}
     }
  \end{center}
     \caption{Visual comparison of the proposed semantic spaces for structuring the VQA class answers: Glove (on the left) and Co-oc (on the right). The embeddings of all classes from the VQAv2-CP dataset in the respective spaces are illustrated via t-SNE~\cite{maaten2008visualizing}. The big circles of various hues represent 4 different categories of the VQAv2-CP answer classes which have been chosen for the sake of illustration. The table lists the different categories, the color in which they are represented in the figure as well as the VQA answer classes of each category.
     }
  \label{fig:grigory_visualization}
\end{figure}

\noindent
\myparagraph{Differences of the spaces Glove and Co-oc} The two embeddings are of different nature, one being estimated from word occurrences in large language corpuses, the second one directly exploiting human annotation. While Co-oc has been estimated in a goal-driven way and, for this reason, arguably could be more adapted to structuring output spaces, we should note that its coverage is smaller. Glove is estimated from large-scale corpuses and it can therefore be expected that any reasonable combination of words has received a significant statistical support during its estimation. On the other hand, the size of the human annotations in the VQA datasets is limited, and equation (\ref{eq:cooc}) estimating the Co-oc vector is dominated by co-occurrences caused by ambiguities in question formulation and human interpretation of questions, content and reasoning. Pairs of answer classes, which are far from each other semantically, therefore receive a small support for statistical estimation making the estimates noisy for these cases. It is difficult to say which one of these two effects --- statistical support and goal-driven nature of estimate --- will take the upper hand. So, this question is answered in the experimental section. 

However, in order to qualitatively confirm that the designed embeddings are correctly encoding the targeted semantic proximity, we show in Figure~\ref{fig:grigory_visualization} the 2D t-SNE projection~\cite{maaten2008visualizing} of Glove and Co-oc embeddings corresponding to the answers classes in the VQAv2-CP dataset~\cite{vqa-cp}. In Figure~\ref{fig:grigory_visualization}, we manually selected four classes’ categories which are semantically close, namely \textit{colors} (10 classes), \textit{dog species} (9 classes), \textit{motorcycle brands} (4 classes) and \textit{tree species} (4 classes). As we can see in the Figure, these categories correspond to spatially grouped clusters in the 2D projection space, both for Glove and Co-oc embeddings, thus confirming their relevance.
\vspace{3mm}

\begin{figure}[t]
  \begin{center}
     \includegraphics[width=.8\linewidth]{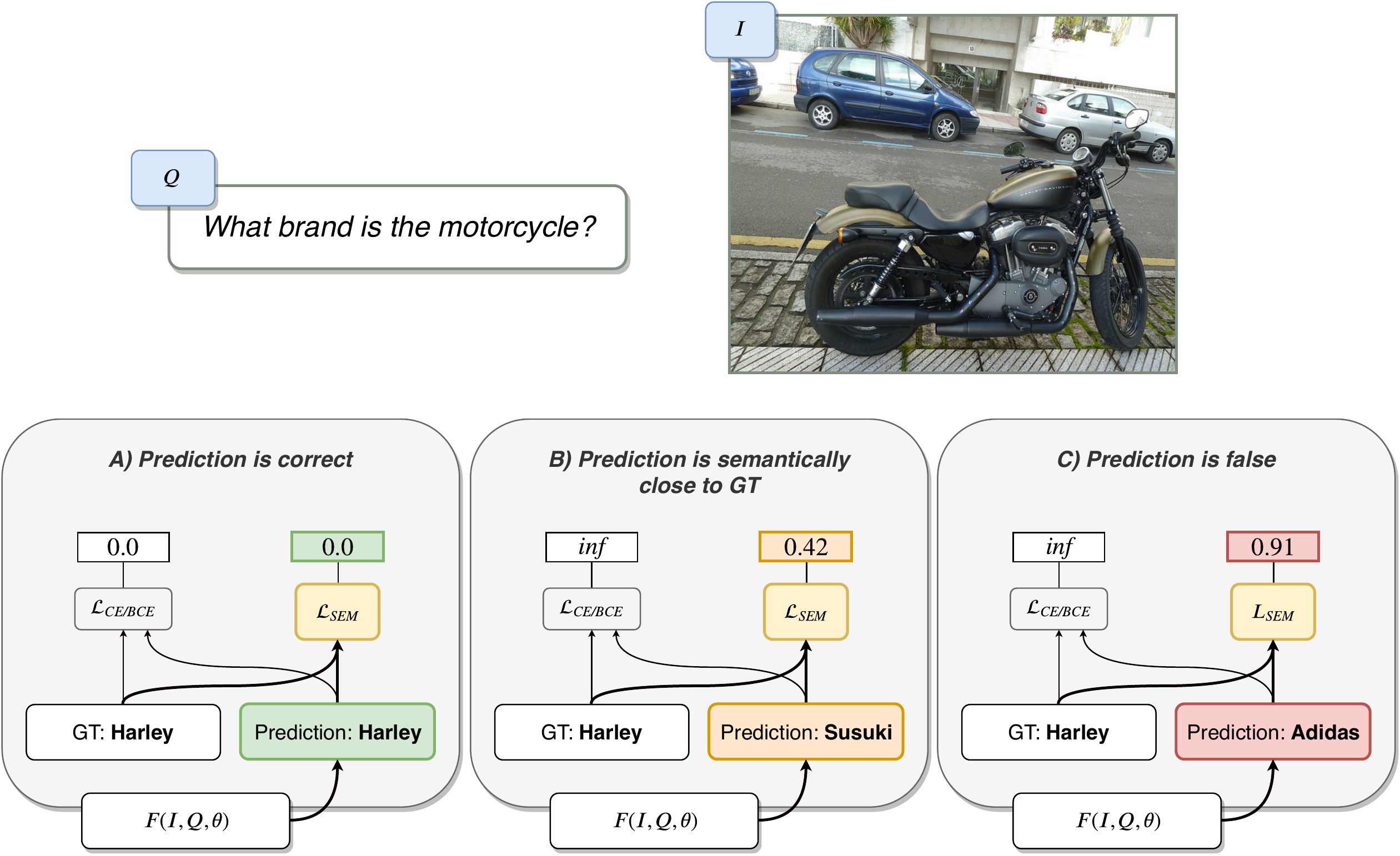}
  \end{center}
     \caption{Schematic illustration of the proposed loss $\mathcal{L}_{\textit{SEM}}$, which penalizes the semantic distance between the prediction and the ground truth answer (GT). When the prediction is false but semantically related to the ground truth -- `\textit{Suzuki}' \textit{vs.} `\textit{Harley}' in case B -- we obtain a lower $\mathcal{L}_{\textit{SEM}}$ value than when the prediction is unrelated to the ground truth -- `\textit{Adidas}' \textit{vs.} `\textit{Harley}' in case C. At the same time, the traditional cross-entropy loss $\mathcal{L}_{\textit{CE/BCE}}$ does not make the difference between case B and C. Numerical values are obtained using the semantic loss with Co-oc semantic space.}
  \label{fig:ex_loss}
\end{figure}

\noindent
\myparagraph{Distances in the semantic space}
\label{subsection:dist} To compensate for differences in normalization, we chose the cosine similarity as a measure for proximity in the embedding spaces:
\begin{equation}
  s(x, y) = \frac{x\cdot y}{||x|| \ ||y||}\textit{,}
\end{equation}
The semantic loss $\mathcal{L}_{\textit{SEM}}$ penalizing miss-classifications between
predictions $\bm{y}$ and targets $\bm{y}^*$ according to semantic proximity is then given as:
\begin{equation}
  \mathcal{L}_{\textit{SEM}} = 1 - s\left(\gamma(\bm{y}), \gamma(\bm{y}^*)\right)\textit{,}
\end{equation}
As described in Figure~\ref{fig:ex_loss}, $\mathcal{L}_{\textit{SEM}}$ allows to take into account the semantic proximity between the prediction and the ground truth, whereas $\mathcal{L}_{\textit{CE/BCE}}$ does not make the difference between wrong predictions which are close or unrelated to the ground truth (respectively case B and C in the Figure~\ref{fig:ex_loss}). Finally, we combine the semantic loss with the classical cross-entropy (or binary cross-entropy) as follows:
\begin{equation}
  L = \mathcal{L}_{\textit{CE/BCE}} + \lambda  \mathcal{L}_{\textit{SEM}}\textit{,}
\end{equation}
where $\lambda$ is a hyper-parameter.

At test time, we remove the semantic loss component and predict an answer as the highest logit/probability.
Our method can thus be applied to any VQA architecture, and only requires the supplementary loss $\mathcal{L}_{\textit{SEM}}$ during the training.

\section{Experiments}
\label{sec:experiments}


We evaluate our contributions on the following two VQA corpuses:
\begin{description}
\item[VQAv2] has been proposed in \cite{goyal2017making} and
 contains open-ended questions in natural language about real images. The
corpus gathers 265K images with at least 3 questions for each. Each question is annotated with 10 ground-truth answers.
\item [VQAv2-CP] has been introduced in ~\cite{vqa-cp}. It has been constructed by reorganizing the training and validation splits of the VQAv2~\cite{goyal2017making} dataset in order to explicitly make the distribution of answers different between the training and test.
In other words, the VQAv2-CP dataset has been designed to measure the sensitivity of a VQA model to the language bias, and therefore is a test for measuring the ability of a model to generalize to unseen situations.
\end{description}
The evaluation of our contribution on VQAv2-CP is a particularly interesting setup, as obtaining good results there requires an agent to reason beyond exploiting biases. This is important in the context of our contribution. While classical auxiliary losses add additional difficulty to a learning task, as for instance self-supervised contrastive losses \cite{OordContrastive2018}, the proposed semantic loss is a different case as it is inherently still based on classification of the answer classes, albeit on a restructured output space. Testing this auxiliary loss in a setting, where the training distribution equals the test distribution would make the evaluation unfavorable by introducing a difference between the minimized objective and the evaluation metric. We claim that the new semantic loss increases reasoning and decreases the dependence on biases, which is better evaluated on datasets with shifts in distributions such as VQAv2-CP. We nevertheless also include comparisons on VQAv2.

To demonstrate that the semantic loss is model-agnostic, we test it on three different standard VQA architectures:
\begin{description}
\item[Bottom-Up-Top-Down (UpDn)]  ~\cite{anderson2018bottom}, is a strong baseline architecture for VQA. It introduced the use of a bottom-up (from pixels to visual objects) visual attention to the standard top-down mechanism. In particular, UpDn uses an object detector --- R-CNN~\cite{ren2015faster} --- to extract bounding boxes along with dense visual features for each object in the image. Thereby, the question attention is computed over a set of objects rather than over standard grid features as has been done before in the literature.
\item[Bilinear Attention Network (BAN)]~\cite{kim2018bilinear} adds a bilinear attention operator on top of the bottom-up top-down mechanism introduced in \cite{anderson2018bottom}. Moreover, this model allows multi-hop reasoning by stacking multiple bilinear attention layers with residual connections.
\item[Deep Modular Co-Attention Networks (MCAN)]~\cite{yu2019deep} is a Transformer-based~\cite{vaswani2017attention}
multi-modal architecture aiming at modeling both the interactions inside one modality (between words or visual objects) and the interactions between the two modalities (between words and objects) using self-attention mechanisms. Like BAN~\cite{yu2019deep}, this architecture contains several stacked self-attention blocks in order to perform complex multi-hop reasoning.
\end{description}

\noindent
We complement our evaluation by showing the complementarity of our method with RUBi~\cite{cadene2019rubi}, a
SOTA approach focusing on the reduction of language bias. It consists of a training procedure adding a
question-only branch with a masking mechanism to the base VQA model during the training. The RUBi module adapts the prediction of the base model in order to prevent it from fully exploiting a language-only bias. At test time, the question-only branch is removed.

\myparagraph{Training details}
We train all our models during 40 epochs.
We use the 6-layers version of MCAN~\cite{yu2019deep}. At the beginning of the training we linearly increase the learning rate from $1e^{-4}$ to $2e^{-1}$ during 8 epochs, followed by a decay by a factor of $0.2$ at epochs 10 and 20. We set the batch size to 64. 
For UpDn~\cite{anderson2018bottom} and BAN~\cite{kim2018bilinear} we set the batch size to 512 and increase the learning rate from $2e^{-3}$ to $2e^{-1}$ during the first 8 epochs, followed by a decay by a factor of $0.2$ at epochs 10 and 20. We use the 4-layers implementation of BAN.
We use binary cross entropy along with the Adam optimizer~\cite{kingma2014adam} for MCAN~\cite{yu2019deep} and Adamax~\cite{kingma2014adam} for BAN~\cite{kim2018bilinear} and UpDn~\cite{anderson2018bottom}\footnote{For the baseline models, we use publicly available implementations at https://github.com/MILVLG/openvqa}.

All of our experiments are run on two NVIDIA P100 GPUs with half precision training using the apex library.\footnote{https://github.com/NVIDIA/apex}
Note that all of our models are trained on the training split only, without the help of any external dataset such as in \cite{kim2018bilinear} and \cite{yu2019deep}.
We set the two hyper-parameters $\lambda=10.0$ and $k=10$ using grid search.

\subsection{Results}

\myparagraph{Model-agnosticity}
Table~\ref{table:model} shows the effectiveness of our semantic loss on VQAv2-CP
with three models. The proposed approach improves  accuracy by $+2.0$, $+0.9$ and
$+0.5$ points on, respectively, MCAN, BAN and UpDn models when using the Glove semantic space. The gain is in general even higher when using Co-oc semantic space, leading to improvements of of $+2.9$ (MCAN), $0.8$ (BAN) and $+0.1$ (UpDn).
The on-par performance of Co-oc with respect to Glove (or arguably superior), despite the estimation of Glove from large-scale datasets, illustrates the strength of the goal-directed estimation strategy of Co-oc and the thus learning signal directly derived from human annotations.

We observe that the impact of the semantic loss on the UpDn architecture is less significant than on BAN and MCAN, both for the Co-oc and Glove semantic spaces.
We conjecture that this is due to the higher dependency of UpDn on the question bias.
To further investigate this, we performed experiments combining the semantic loss with a state-of-the-art bias reduction method.


\setlength{\tabcolsep}{4pt}
\begin{table}[t]
\begin{center}
\caption{Consistency of the performance gains over multiple neural model architectures: performance using MCAN~\cite{yu2019deep}, BAN~\cite{kim2018bilinear} and UpDn~\cite{anderson2018bottom} on VQAv2-CP. Baselines marked with $\dagger$ have been trained by ourselves.}
\label{table:model}
\centering
\begin{tabular}{lclll}
\hline\noalign{\smallskip}
Model & Semantic loss & Embedding & Test Acc. \\
\noalign{\smallskip}
\hline
\noalign{\smallskip}
MCAN~\cite{yu2019deep} $\dagger$ & \xmark &       & 42.5 \\
MCAN & \cmark & Glove & 44.5 \\
MCAN & \cmark & Co-oc & 45.4 \\
\hline
BAN~\cite{kim2018bilinear} $\dagger$ & \xmark &        & 40.6 \\
BAN & \cmark & Glove  & 41.5 \\
BAN & \cmark & Co-oc  & 41.4 \\
\hline
UpDn~\cite{anderson2018bottom} $\dagger$ & \xmark &        & 40.4 \\
UpDn & \cmark & Glove  & 40.9 \\
UpDn & \cmark & Co-oc  & 40.5 \\
\hline
\end{tabular}
\end{center}
\end{table}
\setlength{\tabcolsep}{1.4pt}

\myparagraph{Complementarity of gains with bias-reduction methods}
We combine the semantic loss with RUBi~\cite{cadene2019rubi},
a state-of-the-art method designed to reduce language biases in VQA models. RUBi combines standard VQA models with a second question-only branch, whose objective is the explicit estimation of language biases. During training time, the prediction of the question-only branch is used as a mask combined with the VQA branch by element-wise multiplication, which drives the VQA model to overcome the inherent language bias. The masking is removed during testing.

As shown in Table~\ref{table:rubi}, the semantic loss (in the using Glove variant) improves upon the combination of UpDn model architecture + RUBi training with a margin of $+3.3$ points and reaches an accuracy of $47.5\%$ on the VQAv2-CP test split\footnote{We observe an instability when using UpDn+RUBi which occasionally prevents the model from converging. As a consequence, we provide the average accuracy over four converged models with random seeds along with the standard deviation.}.
This indicates that the proposed loss is complementary to existing bias-reduction approaches and improves the generalization and reasoning abilities of the model.

\setlength{\tabcolsep}{4pt}
\begin{table}[t]
\begin{center}
\caption{Complementary of gains: the semantic loss can be combined with SOTA methods decreasing language biases like RUBi~\cite{cadene2019rubi}, showing combined gains on VQAv2-CP~\cite{vqa-cp}. The semantic space is Glove~\cite{pennington2014glove}. Baselines with $\dagger$ have been trained by us.}
\label{table:rubi}
\begin{tabular}{ll}
\hline\noalign{\smallskip}
Model & Test Acc. \\
\noalign{\smallskip}
\hline
\noalign{\smallskip}
UpDn~\cite{anderson2018bottom} $\dagger$ & 40.4 \\
+ RUBi~\cite{cadene2019rubi} & 44.23 \\
+ RUBi~\cite{cadene2019rubi} + Semantic Loss (ours) & 47.5 $\pm$ 0.3 \\
\hline
\end{tabular}
\end{center}
\end{table}
\setlength{\tabcolsep}{1.4pt}

\setlength{\tabcolsep}{4pt}
\begin{table}[t!]
\begin{center}
\caption{Comparison on VQAv2~\cite{goyal2017making}. Among the work focusing on reducing language biases, the proposed semantic loss is the method which degrades performance the least on VQAv2. *For a fair comparison, we display the accuracy of the base model used by the authors of the different methods. We only display models which does not rely on additional annotation such as \cite{selvaraju2019taking} and \cite{wu2019self}. Baselines with $\dagger$ have been trained by us.}
\label{table:vqa}
\begin{tabular}{lcc}
\hline\noalign{\smallskip}
Model & VQAv2-val & $\Delta$ \\
\noalign{\smallskip}
\hline
\noalign{\smallskip}
UpDn* & 62.9 & \\
UpDn + DLR~\cite{jingovercoming} & 58.0 & -4.9\\
\hline
Base* & 63.1 & \\
Base + RUBi~\cite{cadene2019rubi} & 61.2 & -2.6\\
\hline
UpDn* & 63.5 & \\
UpDn + SCR (QA) ~\cite{wu2019self} & 62.3 & -1.2\\
\hline
UpDn* & 63.5 & \\
UpDn + Q-Adv + DoE~\cite{ramakrishnan2018overcoming} & 62.8 & -0.7\\
\hline
MCAN $\dagger$ & 66.1 & \\
\textbf{MCAN + Semantic Loss (ours)} & \textbf{66.0} & \textbf{-0.1}\\
\hline
\end{tabular}
\end{center}
\end{table}
\setlength{\tabcolsep}{1.4pt}

\setlength{\tabcolsep}{4pt}
\begin{table}[t!]
\begin{center}
\caption{Comparison of our method combined with RUBi~\cite{cadene2019rubi} against state-of-the-art on VQAv2-CP. We divide the table in two groups: methods based on the UpDn~\cite{anderson2018bottom} architecture -- on the bottom part -- and the others -- on the top part. Column `Supp. ann.' specifies models trained with additionnal annotations. Models with $\dagger$ have been trained by us. Base* correspond to the baseline model used in \cite{cadene2019rubi}}
\label{table:sota}
\begin{tabular}{lcc}
\hline\noalign{\smallskip}
Model & Test Acc. & Supp. ann.\\
\noalign{\smallskip}
\hline
\noalign{\smallskip}
%
Question-Only~\cite{vqa-cp} & 15.95 &\\
BAN~\cite{kim2018bilinear} $\dagger$ & 40.6 &\\
Q-type Balanced Sampling~\cite{cadene2019rubi} & 42.1 &\\
MCAN~\cite{yu2019deep} $\dagger$ & 42.5 &\\
NSM~\cite{hudson2019learning} & 45.8 &\\
Base* + RUBi~\cite{cadene2019rubi} & 47.1 &\\
\hline
UpDn~\cite{anderson2018bottom} $\dagger$ & 40.4 &\\
UpDn + Q-Adv + DoE~\cite{ramakrishnan2018overcoming} & 41.2 &\\
UpDn + RUBi~\cite{cadene2019rubi} & 44.2 &\\
UpDn + HINT~\cite{selvaraju2019taking} & 46.7 & \cmark\\
\textbf{UpDn + RUBi + Semantic Loss (ours)} & \textbf{47.5} &\\
UpDn + SCR (QA)~\cite{wu2019self} & 48.5 &\\
UpDn + DLR~\cite{jingovercoming} & 48.9 &\\
UpDn + SCR (VQA-X)~\cite{wu2019self} & 49.5 & \cmark\\
\hline

\end{tabular}
\end{center}
\end{table}
\setlength{\tabcolsep}{1.4pt}

\myparagraph{Impact on VQAv2 dataset} As discussed before, the VQAv2-CP dataset has been proposed by ~\cite{ramakrishnan2018overcoming} with the goal to evaluate the performance of VQA models in a condition where they cannot fully rely on question biases.
Indeed, as shown in \cite{ramakrishnan2018overcoming}, the original VQAv2~\cite{goyal2017making} dataset contains numerous question biases (\textit{e.g.} the question \say{what color is the banana in the picture} can be correctly answered as \say{yellow} without even analyzing the picture in VQAv2).
At the same time, it is very important to verify that our semantic loss, which is effective for training VQA models on the \textit{unbiased} VQAv2-CP dataset, does not hinder the model's performances when the training is done on the \textit{biased} VQAv2 dataset.

Therefore, Table~\ref{table:vqa} analyzes the impact of our semantic loss on the VQAv2 dataset and compares it with recent approaches designed to remove the language bias in VQA.
More precisely, we compare the accuracies on the VQAv2 validation dataset of baseline VQA models (the original baselines from the respective works are taken) with and without one of the SOTA approaches aiming to reduce the question biases.
For a fair comparison, we only compare with methods which does not rely on extra annotated supervision such as HINT~\cite{selvaraju2019taking} and SCR(VQA-X)~\cite{wu2019self}.
The impacts of the compared approaches on the respective baselines are highlighted in the \say{$\Delta$} column of Table~\ref{table:vqa}.

When combining the semantic loss  with MCAN~\cite{yu2019deep} in the Co-oc variant, we observe a marginal drop of $-0.1$ in accuracy.
On the contrary, SCR (QA)~\cite{wu2019self} and RUBi~\cite{cadene2019rubi} cause significant drops of performance of respectively $-1.2$ and $-2.6$ points.
The drop of the recent DLR~\cite{jingovercoming} method is even more impressive reaching almost $-5$ accuracy points.
All in all, contrary to other SOTA methods, our semantic loss allows to reduce the dependency on question biases (\textit{cf.} results presented in Table~\ref{table:model}) without sacrificing accuracy on the biased VQAv2 dataset.

\myparagraph{Comparison with the state-of-the-art}
We compare our method with SOTA approaches on the VQAv2-CP dataset in Table~\ref{table:sota}. For a fair comparison, we divide Table~\ref{table:sota} into two groups: methods based on the UpDn~\cite{anderson2018bottom} architecture and the others. When combining the proposed loss along with another bias reduction method -- namely RUBi~\cite{cadene2019rubi} -- using the UpDn architecture, we achieve a SOTA-level accuracy of $47.5\%$.  
Note that, contrary to HINT~\cite{selvaraju2019taking} or SCR(VQA-X)~\cite{wu2019self}, our approach does not require any additional annotations.
DLR~\cite{jingovercoming} achieves a high accuracy ($48.5\%$) on VQAv2-CP. However, unlike our approach, DLR causes a significant drop in accuracy (of $-4.9$ points) on the biased VQAv2 dataset as highlighted in Table~\ref{table:vqa}.



\section{Conclusions}
VQA as almost always been treated as a classification task. However, despite its convenience, this strategy does not take into account the semantic relationships between the answers.
We showed that suitably structuring the semantic space of output classes can overcome some of the shortcomings of the classification strategy widely used in VQA.
We propose a new loss based on proximity in a semantic space and we 
suggest two different ways to estimate  semantic proximity. One is based on word embeddings, the second one directly taps into human assessments and exploits the ambiguities of question formulations and their interpretation. We show that while this proximity space is estimated from  less statistical coverage, it is not less effective.

We experimentally demonstrate the effectiveness of the semantic loss in reducing dependency over language biases on VQAv2-CP~\cite{vqa-cp} and its consistency over several standard VQA architectures. Moreover, we show that contrary to other SOTA methods, this gain in accuracy is not made at a cost of a degradation of the performance on the classic VQAv2~\cite{goyal2017making} dataset.
Finally, when combined with another bias reduction method, our semantic loss allows to achieve an accuracy on par with SOTA on VQAv2-CP~\cite{vqa-cp}.

In future work, we aim to continue to pave the way to the long-term objective of the community in moving to a direct structured prediction of the textual output, which will bring VQA closer to more traditional models in NLP. Even in this case of direct prediction of text sequences, however, it is far from clear whether classification as an auxiliary loss could not eventually provide an additional useful and complementary learning signal.
\clearpage
%
%
\bibliographystyle{unsrt}
\bibliography{egbib}
\end{document}